\documentclass[conference]{IEEEtran}
\IEEEoverridecommandlockouts
\usepackage{cite}
\usepackage{amsmath,amssymb,amsfonts}
\usepackage{algorithmic}
\usepackage{graphicx}
\usepackage{textcomp}
\usepackage{xcolor}
\usepackage{hyperref}
\usepackage{float}
\usepackage{algorithm}
\usepackage{algorithmic}
\usepackage{xcolor}
\usepackage{comment}
\usepackage{subcaption}
\usepackage{tikz}
\usepackage{stfloats}
\usepackage{booktabs}
\usepackage{multirow}

\usetikzlibrary{shapes.geometric, arrows.meta, positioning, fit, backgrounds, calc, shadows}
\usepackage{amsmath}
\usepackage{amssymb}
\usepackage{caption} 

\hypersetup{hidelinks}

\def\BibTeX{{\rm B\kern-.05em{\sc i\kern-.025em b}\kern-.08em
    T\kern-.1667em\lower.7ex\hbox{E}\kern-.125emX}}
\begin{document}

\title{Adaptive Entropy-Driven Sensor Selection in a  Camera-LiDAR Particle Filter for Single-Vessel Tracking\\}
\author{%
\IEEEauthorblockN{%
\IEEEauthorrefmark{1}Andrei Starodubov,
\IEEEauthorrefmark{2}Yaqub Aris Prabowo,
\IEEEauthorrefmark{1}Andreas Hadjipieris,
\IEEEauthorrefmark{1}Ioannis Kyriakides,
\IEEEauthorrefmark{2}Roberto Galeazzi
}
\IEEEauthorblockA{
\href{mailto:andrei.starodubov@cmmi.blue}{andrei.starodubov@cmmi.blue}, 
\href{mailto:yaqpr@dtu.dk}{yaqpr@dtu.dk},
\href{mailto:andreas.hadjipieris@cmmi.blue}{andreas.hadjipieris@cmmi.blue},
\href{mailto:ioannis.kyriakides@cmmi.blue}{ioannis.kyriakides@cmmi.blue},
\href{mailto:roga@dtu.dk}{roga@dtu.dk}
}
\IEEEauthorblockA{\IEEEauthorrefmark{1}%
Cyprus Marine and Maritime Institute, Larnaca, Cyprus\
} 
\IEEEauthorblockA{\IEEEauthorrefmark{2}%
Technical University of Denmark, Lyngby, Denmark\
}
}

\maketitle

\begin{abstract}
Robust single-vessel tracking from fixed coastal platforms is hindered by modality-specific degradations: cameras suffer from illumination and visual clutter, while LiDAR performance drops with range and intermittent returns. We present a particle-filter tracker that supports sequential measurement-level camera–LiDAR fusion and an information-gain (entropy-reduction) adaptive sensing policy that selects the most informative sensing modality at each fusion time bin. The approach is validated in a real maritime deployment at the \textit{Cyprus Marine and Maritime Institute Smart Marina Testbed} (Ayia Napa Marina, Cyprus), using a shore-mounted 3D LiDAR and an elevated fixed camera to track a rigid inflatable boat with onboard GNSS ground truth. We compare LiDAR-only, camera-only, All sensors, and adaptive configurations. Results show LiDAR dominates near-field accuracy, the camera sustains longer-range coverage when LiDAR becomes unavailable, and the adaptive policy achieves a favorable accuracy–continuity trade-off by switching modalities based on information gain. The adaptive configuration therefore provides a practical sensor-selection baseline for resilient and resource-aware maritime surveillance. 
\end{abstract}


\section{Introduction}
Persistent and accurate tracking of surface vessels is fundamental to maritime security, traffic monitoring, and port management. Robust tracking systems enable critical applications from collision avoidance to sovereignty enforcement. However, achieving reliable performance in the variable and often hostile marine environment remains an unsolved challenge. A primary obstacle is the inherent physical limitation of any single sensing modality, which restricts system reliability under the full spectrum of operational conditions.

Optical cameras offer rich visual data and are well established for maritime tracking, with methods from background subtraction to deep learning enabling vessel identification and classification \cite{kim2014background,chen2020robust}. However, vision systems are constrained by illumination and weather-related occlusion, becoming unreliable at night or in precipitation. By contrast, LiDAR provides precise, light-invariant 3D geometry for object detection and motion tracking \cite{qi2023real}, yet its performance degrades with distance, fog, and heavy rain, and it conveys limited appearance-based identification information.

Related work relevant to the present problem spans several in directions. Camera-only approaches estimate vessel motion from monocular imagery using appearance cues and scene geometry, including visual ship tracking and metric monocular tracking with GNSS-based calibration \cite{chen2020robust,Tobias_marine_track_monocular_camera_2021}. LiDAR-only maritime pipelines have also been reported, from real-time vessel detection and tracking to LiDAR-based ship tracking in busy maritime environments \cite{qi2023real, XIE_lidar_detection_2024}. Beyond single-modality methods, heterogeneous maritime tracking has been addressed with multi-target and track-fusion architectures that combine camera, LiDAR, radar, and related sensors \cite{helgesen2022asv_heterogeneous,han2019comparison}. Recent marine camera–LiDAR systems further study association and cluster-selection mechanisms for combining LiDAR geometry with visual discrimination in near-shore scenes \cite{dalhaug2025near, OBRADOVIC_LiDAR_Camera_Fusion_2024}. To mitigate the weaknesses of individual sensors, multi-sensor fusion has emerged as a promising direction. In maritime contexts, prior work has explored fusion primarily between two modalities, such as camera with LiDAR \cite{dalhaug2025near}. Still, robust tracking from a fixed observation platform under widely varying conditions remains challenging, particularly when one modality becomes temporarily unreliable.

This paper addresses this need by proposing a particle-filter tracker for robust single-vessel tracking from a fixed coastal platform using a shore-mounted 3D LiDAR and an elevated fixed camera. The method supports sequential measurement-level fusion of the available sensor likelihoods when both modalities are used, while also allowing single-modality operation when only one sensor is used or preferred. In addition, we introduce an adaptive policy inspired by information-theoretic sensor management\cite{kyriakides2021agile, Xu_Gaussian_Process_Based_EIG}. At each fusion time bin, the predicted state estimate is used to form hypothetical updates for the candidate sensing modalities, and the modality that yields the largest expected entropy reduction is selected for the actual update. 

The remainder of this paper is structured as follows: Section II formulates the tracking problem. Section III details our sensor configuration and setup. Sections IV and  V present our methodology, including data preprocessing, a multi-stage boat detection pipeline for each sensor, and the architecture of our fused particle filter. Section VI reports experimental results and provides a comparative analysis. Finally, Section VII concludes the paper and discusses avenues for future work.
\label{sec:Introduction}

\section{Problem Formulation}
The tracking problem is formulated within a Bayesian filtering framework, where the objective is to recursively estimate the state of a vessel $\mathbf{x}_k \in \mathbb{R}^4$ at discrete time step $k$, given all available sensor measurements $\mathbf{z}_{1:k}$ up to that time.

\subsection{State Vector}
The vessel state is defined as:
\begin{equation}
\mathbf{x}_k = \left[ p_x,\ p_y,\ v_x,\ v_y \right]^\top \in \mathbb{R}^4
\label{eq:state_vector}
\end{equation}
where:
\begin{itemize}
    \item $p_x, p_y \in \mathbb{R}$: 2D Cartesian coordinates in a world coordinate system.
    \item $v_x, v_y \in \mathbb{R}$: Linear velocities along the $x$ and $y$ axes, respectively.
\end{itemize}

\subsection{Vessel Motion Model}
\label{subsec:motion_model}
The vessel dynamics follows a linear constant velocity (CV) model:
\begin{equation}
\mathbf{x}_k = f(\mathbf{x}_{k-1}) + \mathbf{\epsilon}_k,\quad \mathbf{\epsilon}_k \sim \mathcal{N}(0, \mathbf{Q})
\label{eq:motion_model}
\end{equation}
where $\mathbf{x}_k = [p_{x,k}, p_{y,k}, v_{x,k}, v_{y,k}]^\top \in \mathbb{R}^4$ represents the position and Cartesian velocities. The term $\mathbf{\epsilon}_k \in \mathbb{R}^4$ represents unknown disturbances modeled as zero-mean Gaussian noise with covariance $\mathbf{Q} \in \mathbb{R}^{4 \times 4}$.

The state transition function $f: \mathbb{R}^4 \rightarrow \mathbb{R}^4$ is:
\begin{equation}
f(\mathbf{x}_{k-1}) = 
\begin{bmatrix}
p_{x,k-1} + v_{x,k-1}\Delta t \\
p_{y,k-1} + v_{y,k-1}\Delta t \\
v_{x,k-1} \\
v_{y,k-1}
\end{bmatrix}
\label{eq:state_transition}
\end{equation}
where $\Delta t \in \mathbb{R}^+$ is the sampling interval.

\subsection{Observation Models}
The system employs 
two 
heterogeneous sensors, each providing measurements $\mathbf{z}_k^{(s)}$ at potentially different rates. The composite observation model is given by:
\begin{equation}
\mathbf{z}_k = \left\{ \mathbf{z}_k^{\text{(cam)}},\ \mathbf{z}_k^{\text{(lidar)}}\ 
\right\}
\label{eq:composite_observation}
\end{equation}

\subsubsection{Camera}
The camera provides image-plane measurements after foreground extraction:
\begin{equation}
\mathbf{z}_k^{\text{(cam)}} = [u_k,v_k]^\top \in \mathbb{R}^2
\label{eq:cam_obs}
\end{equation}
where $(u_k,v_k) \in \mathbb{R}^2$ are the pixel coordinates of the bounding box bottom-center.

\subsubsection{LiDAR}
The 3D LiDAR provides a set of $N_l$ points from the detected vessel cluster in the sensor's polar coordinate system:
\begin{equation}
\mathbf{z}_k^{\text{(lidar)}} = \left\{ (r_j,\ \theta_j) \mid j = 1,\dots,N_l \right\},\quad (r_j, \theta_j) \in \mathbb{R}^2
\label{eq:lidar_obs}
\end{equation}
where $r_j \in \mathbb{R}^+$ and $\theta_j \in (-\pi, \pi]$ are the range and bearing, respectively, of the $j$-th point relative to the LiDAR's fixed position.





\label{sec:Problem_Formulation}

\section{Sensor Configuration}
\begin{figure}[!b]
\centering
\includegraphics[width=0.99\columnwidth]{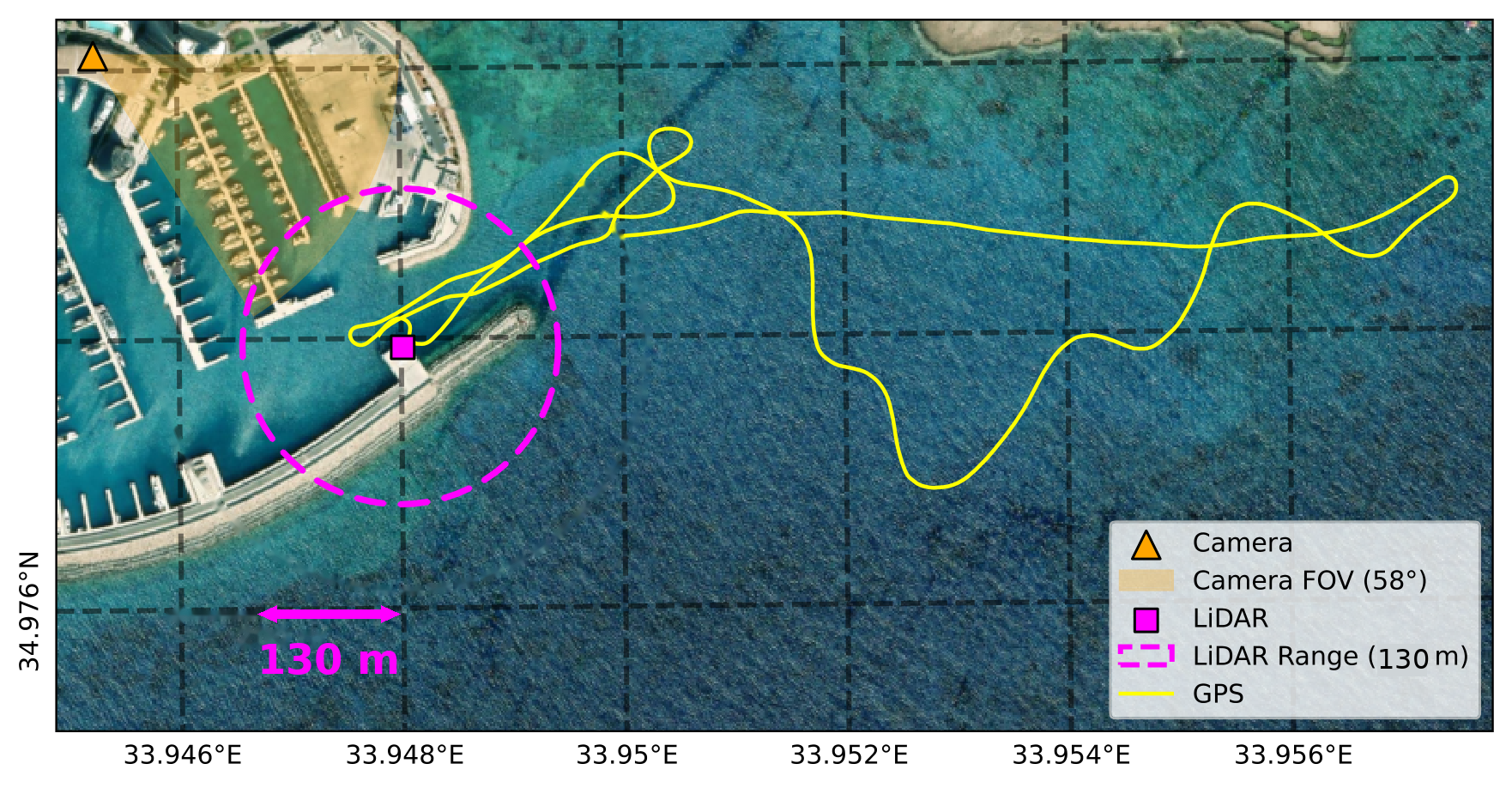}
\caption{Multi-sensor platform (Camera and LiDAR) deployed at a coastal location at Ayia Napa Marina.}
\label{fig:coordinate_frames}
\end{figure}

\begin{figure}[!t]
\centering
\includegraphics[width=0.99\columnwidth]{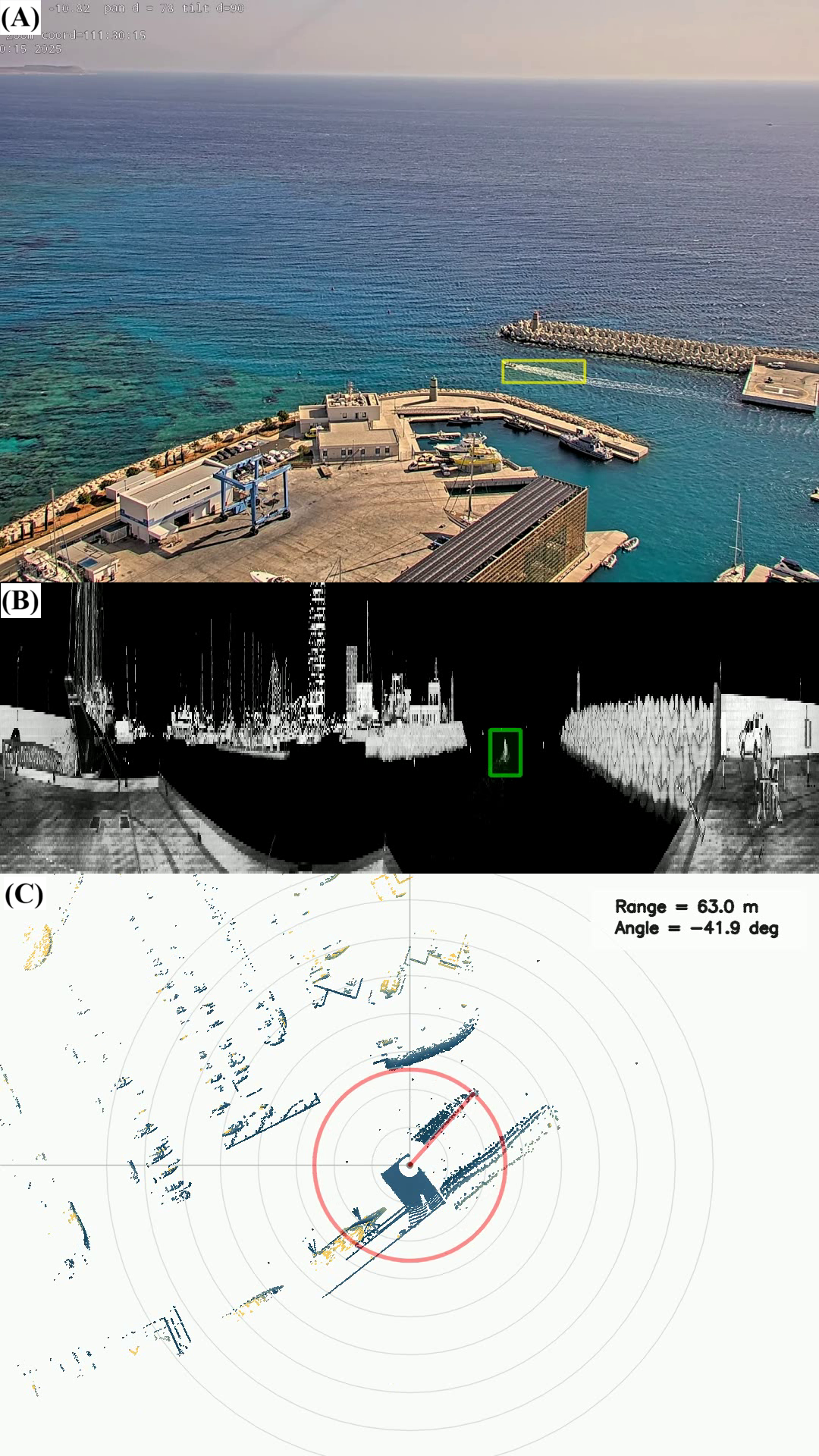}






\caption{(A) Camera frame with detected bounding box. (B) LiDAR visualization: a spherical intensity map with a bounding box overlay. (C) LiDAR visualization: a Bird’s Eye View plotting the target's specific range radius and azimuth bearing. 
}
\label{fig:sensor_data}
\end{figure}

Experiments were conducted using the \textit{Cyprus Marine and Maritime Institute (CMMI) Smart Marina Testbed} located at Ayia Napa Marina, Cyprus. The experimental setup consists of a stationary multi-sensor platform as illustrated in Fig. \ref{fig:coordinate_frames}. Fig. \ref{fig:sensor_data} shows simultaneous data from 
both sensors during a boat passage.

A calibrated AXIS Q8752-E camera (1920x1080 resolution, 12-15 fps, $58^{\circ}$ FOV) is mounted 100 m above sea level with a direct view of the waterway. An Ouster OS2 LiDAR (130 m range in current testbed, $360^{\circ}$ horizontal, $22.5^{\circ}$ vertical FOV, 10 Hz scan rate) is mounted on the helipad. Its coordinate system is transformed to align with the camera's world frame. 

\label{sec:Sensor_Configuration}

\section{Calibration and Detection}
\begin{figure*}[!b]
\centering
\resizebox{\textwidth}{!}{\usetikzlibrary{arrows.meta,positioning,calc,fit,matrix,shapes.geometric}

\begin{tikzpicture}[
    node distance=1.5cm and 2cm,
    font=\sffamily\small,
    sensor/.style={
        draw, ellipse, 
        fill=blue!10, 
        minimum height=1cm, 
        minimum width=3.5cm, 
        align=center
    },
    process/.style={
        draw, rectangle, rounded corners, 
        fill=white, 
        minimum height=1.2cm, 
        text width=4cm, 
        align=center,
        drop shadow
    },
    filter/.style={
        draw, rectangle, 
        fill=orange!10, 
        minimum height=6cm, 
        minimum width=8.6cm, 
        align=left, 
        rounded corners
    },
    output/.style={
        draw, trapezium, trapezium left angle=70, trapezium right angle=110,
        fill=green!10, 
        minimum height=1cm, 
        align=center
    },
    line/.style={-Latex, thick}
]

    \node[sensor] (cam) {Camera \\ (RGB)};
    \node[sensor, below=of cam, yshift=1cm] (lidar) {LiDAR \\ (3D Points)};

    
    \node[process, right=of cam] (proc_cam) {
        \textbf{Detection Pipeline}\\
        MOG2 \& BBox\\
        \textit{Homography } $\mathbf{H}_{\mathcal{C}\to\mathcal{W}}$
    };

    \node[process, right=of lidar] (proc_lidar) {
        \textbf{Detection Pipeline}\\
        MOG2, Clustering \& Filtering\\
        \textit{Transform } $\mathbf{R}, \mathbf{t}$
    };


    \node[filter, right=of proc_lidar, xshift=0cm] (pf) {};
    
    \node[below=0.2cm of pf.north] (pf_title) {\textbf{Multi-Modal Particle Filter}};
    
    \node[below=0.3cm of pf_title, align=left, text width=8.4cm, anchor=north] (pf_steps) {
        \textbf{1. Prediction} \\
        \hspace{1em} Motion CV Model: $\mathbf{x}_k = f(\mathbf{x}_{k-1}) + \mathbf{\epsilon}_k$
        \\ [0.2cm]
        \textbf{2. Update} \\
        \hspace{1em} Weights $w_k^{(i)}$ via selected likelihood set: \\
        \hspace{1em} $$\prod_{s\in S_k}^{}L_s$$
        \\ [0.3cm]
        \hspace{1em} where $S_k\in \left\{ \left\{ cam \right\}, \left\{ lidar \right\},\left\{ cam, lidar \right\},\left\{ adaptive\right\} \right\}$
        \\ [0.2cm]
        \textbf{3. Resampling} \\
        \hspace{1em} Based on $N_{\text{eff}}$
    };

    \node[output, right=of pf] (state) {State Estimate \\ $\hat{\mathbf{x}}_k = [p_x, p_y, v_x, v_y]^\top$};

    
    \draw[line] (cam) -- (proc_cam);
    \draw[line] (lidar) -- (proc_lidar);

    \draw[line] (proc_cam.east) -- node[above, font=\footnotesize, pos=0.5] {$\mathbf{z}^{\text{(cam)}}$} (proc_cam.east -| pf.west);
    \draw[line] (proc_lidar.east) -- node[above, font=\footnotesize, pos=0.5] {$\mathbf{z}^{\text{(lidar)}}$} (proc_lidar.east -| pf.west);

    \draw[line] (pf) -- (state);

    \draw[line, dashed] (state.south) -- ++(0,-2.9cm) -| node[pos=0.25, below] {Previous State $\mathbf{x}_{k-1}$} (pf.south);

\end{tikzpicture}}
\caption{System overview showing parallel camera and LiDAR processing streams feeding a particle filter that supports either sequential measurement-level fusion of the available sensor likelihoods or adaptive single-modality selection.}
\label{fig:system_overview}
\end{figure*}
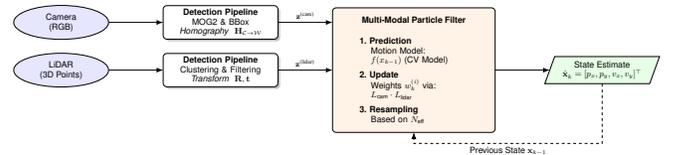

The proposed system operates in two main stages: (1) sensor-specific data preprocessing and detection, and (2) multi-modal fusion using a particle filter. Fig. \ref{fig:system_overview} illustrates the complete processing pipeline.

\subsection{Spatial Calibration to the World Frame}
\label{subsec:preprocessing}

All detections are fused in a common 2D world frame $\mathcal{W}$ defined on the sea-surface plane.
GNSS coordinates (WGS-84 latitude/longitude) are projected to a local metric map using a UTM
Easting/Northing representation. The camera image frame is denoted by $\mathcal{C}$ with pixel
coordinates $\mathbf{u}=[u,\,v]^\top$. The LiDAR scan $\mathcal{L}$ follows the standard convention
($x$ forward, $y$ left, $z$ up), in meters.

Since the camera and LiDAR are static, spatial calibration is performed once offline and reused for
the entire sequence. Under the planar water-surface assumption, the mapping from camera frame pixels to
world-plane coordinates is modeled by a planar homography
$\mathbf{H}_{\mathcal{C}\rightarrow\mathcal{W}}\in\mathbb{R}^{3\times 3}$ \cite{hartley_zisserman_mvg}.
The homography is estimated from  manually selected pixel--world correspondences using RANSAC
for outlier rejection \cite{fischler_bolles_ransac} and is computed using OpenCV with the RANSAC option \cite{opencv_library}. For a
pixel measurement $\mathbf{u}_i=[u_i,\,v_i]^\top$, the corresponding world-plane point
$\mathbf{p}_i=[E_i,\,N_i]^\top$ is obtained as
\begin{equation}
\begin{aligned}
&\mathbf{p}_i
= \pi\!\left(\mathbf{H}_{\mathcal{C}\rightarrow\mathcal{W}}
[u_i,\,v_i,\,1]^\top\right),\\
&\pi([\tilde{E},\tilde{N},\tilde{s}]^\top)
= [\tilde{E}/\tilde{s},\,\tilde{N}/\tilde{s}]^\top .
\end{aligned}
\label{eq:homography}
\end{equation}
where $\pi(\cdot)$ denotes dehomogenization (projective division), and $\tilde{s}\neq 0$ is the
homogeneous scale component.

LiDAR detections are produced in the LiDAR scan $\mathcal{L}$ and mapped to $\mathcal{W}$ via a 2D
planar transform,
\begin{equation}
\mathbf{p}_{\mathcal{W}} = \kappa\,\mathbf{R}(\theta)\,\mathbf{p}_{\mathcal{L}} + \mathbf{t},
\label{eq:lidar_to_world}
\end{equation}
where $\mathbf{p}_{\mathcal{L}}=[x,\,y]^\top$ is the horizontal (sea-plane) LiDAR coordinate,
$\mathbf{R}(\theta)$ is a 2D rotation matrix, and $\mathbf{t}$ is a 2D translation. Setting
$\kappa=1$ yields a rigid SE(2) transform, while allowing $\kappa\neq 1$ yields a 2D similarity
transform. The parameters are estimated from LiDAR/world correspondences using RANSAC for outlier
rejection \cite{fischler_bolles_ransac}, followed by a least-squares refinement using Umeyama's
closed-form alignment \cite{umeyama1991}.

\subsection{Temporal Calibration (GNSS-Camera-LiDAR)}
\label{subsec:temporal_calibration}
The three data streams are not hardware-synchronized and exhibit dropped samples and clock drift. We therefore aligned each sensor clock to a common reference timeline (GNSS time) using manually selected \emph{anchor events} where a GNSS point, a camera frame, and a LiDAR scan correspond to the same physical scene moment.

Let $\{(t^{(k)}_s,\,t^{(k)}_g)\}_{k=1}^{K}$ denote $K$ anchor pairs between a sensor clock $t_s$ (camera or LiDAR) and GNSS time $t_g$. Between two consecutive anchors $k$ and $k{+}1$, the time mapping is modeled as a piecewise affine warp:
\begin{equation}
t_g = a_k\,t_s + b_k,\quad t_s\in[t_s^{(k)},\,t_s^{(k+1)}],
\label{eq:time_warp}
\end{equation}
with
\[
a_k=\frac{t_g^{(k+1)}-t_g^{(k)}}{t_s^{(k+1)}-t_s^{(k)}},
\qquad
b_k=t_g^{(k)}-a_k\,t_s^{(k)}.
\]
After warping, all measurements are placed onto a shared timeline and grouped into fixed-duration bins $\Delta t_{\text{bin}}$. For each sensor and bin, the last available measurement in the bin is retained, yielding a synchronized per-bin set of observations for downstream fusion.

\subsection{Camera Detection Pipeline}
\label{subsec:Camera_Detection_Pipeline}
We detect motion in the RGB stream using background subtraction restricted to the sea region.
A static binary sea mask $\mathbf{M}_{\text{sea}}$ (excluding land and static infrastructure) is applied to all frames.
For each frame $\mathcal{C}_k$, we compute a foreground mask using OpenCV's Mixture of Gaussians (MOG2) background subtraction method.
MOG2 maintains, for each pixel, an adaptive mixture of Gaussians whose parameters are updated online; a pixel is labeled as
foreground if it is not well explained by the current background model \cite{zivkovic2004,zivkovic2006}.

We form the final motion mask by applying the sea ROI and post-processing:
\begin{equation}
\mathbf{M}_k = \mathbb{I}\!\left(\texttt{MOG2}(\mathcal{C}_k) > \tau\right)\ \wedge\ \mathbf{M}_{\text{sea}},
\end{equation}
where $\mathbb{I}(\cdot)$ is an indicator function and $\tau$ is a high threshold (used to obtain a binary foreground mask).
We then apply morphological opening and dilation to remove speckle noise and fill small gaps, extract connected
components/contours, and convert them to axis-aligned bounding boxes.

Background subtraction over maritime scenes can produce spurious foreground blobs caused by wave motion, specular glints, foam, and camera noise. After generating candidate camera bounding boxes from the MOG2 foreground mask and morphological cleanup, we apply a second-stage appearance-based classifier to suppress false positives before downstream processing. For each candidate box, we crop the corresponding image region, resize it to a fixed window of $64\times 64$ pixels, convert it to grayscale, and compute a Histogram of Oriented Gradients (HOG) descriptor (cell size $8\times8$, block size $16\times16$, block stride $8\times8$, $9$ orientation bins). A linear Support Vector Machine (SVM) is trained on approximately $300$ labeled examples to classify each proposal as boat vs non-boat.

To create training labels, candidate proposals are matched against available ground-truth boat annotations using an intersection over union (IoU)-based rule: proposals exceeding a positive IoU threshold are treated as positives, while proposals below a negative IoU threshold are treated as negatives; ambiguous overlaps are ignored. To balance training data, at most a fixed ratio of negatives per positive ($3:1$) is retained. The final classifier is implemented as a standardized feature pipeline (zero-mean/unit-variance normalization) followed by a linear SVM with class-weight balancing ($C=4$, max iterations $6000$). At inference time, each proposal is assigned a linear-SVM decision score $d_j$. Proposals with $d_j > 0.01$ are retained as boat candidates. In our dataset, this denoising stage removed $11634$ out of $26316$ camera proposals ($44.2\%$), substantially reducing false positives and producing a cleaner measurement stream for subsequent fusion.

For each retained merged detection $j$, we assign a heuristic confidence score $c_j\in[0, 1]$ based on the size of the bounding box and the fraction of foreground pixels it contains. Larger and denser motion blobs are treated as more reliable and therefore receive higher confidence, whereas small or sparse blobs receive lower confidence. Thus, the SVM decision score $d_j$ is used only to retain or reject candidate detections whereas the confidence score $c_j$ is passed to the tracker and used as separate motion-based reliability measure.

The image measurements are the bounding-box bottom-center $\mathbf{u}_k=[u_k,v_k]^\top$, which is mapped to the world
frame via the homography in~\eqref{eq:homography}.

\subsection{LiDAR Detection Pipeline}
\label{subsec:LiDAR_Detection_Pipeline}
The Ouster OS2-128 provides per-beam channel fields including \emph{range} (distance-to-return, reported in millimeters),
\emph{signal} (returned signal intensity), \emph{calibrated reflectivity} (a range- and sensitivity-compensated intensity
proxy related to target reflectance), and \emph{near-infrared} (ambient illumination). We primarily use the calibrated reflectivity and range fields.

Each LiDAR scan is first pre-filtered to suppress non-water and spurious returns using conservative range/height/intensity
thresholds. The scan is then represented in the sensor's native $H\times W$ angular grid and destaggered so that columns
correspond to azimuth angle. In our configuration, this yields a reflectivity image of size
$128 \times 2048$ (128 vertical channels and 2048 azimuth samples).

Motion is extracted by applying OpenCV's MOG2 background subtraction to the destaggered reflectivity image
\cite{zivkovic2004,zivkovic2006}. To improve sensitivity to targets moving predominantly in the radial
direction (toward/away from the sensor), we additionally compute a range-change mask from the destaggered range image, e.g.,
by thresholding the absolute difference between the current range image and a slowly varying background range model.
The final foreground mask is formed by combining (logical OR) the reflectivity-based and range-change masks, followed by
morphological cleanup. Connected components/contours are converted to bounding boxes.

Given the set of 3D points inside the  bounding box, we compute a compact polar measurement for tracking. The range
estimate is taken as the median of point ranges (robust to outliers and partial occlusions), and the bearing estimate is taken
as a circular mean of point azimuth angles (robust to angle wrap-around at $\pm\pi$). These summary statistics correspond to
the implementation:
\begin{equation}
\begin{aligned}
\hat{r}_k      &= \mathrm{median}(\{r_j\}),\\
\hat{\theta}_k &= \mathrm{atan2}\!\left(\sum_j \sin\theta_j,\ \sum_j \cos\theta_j\right).
\end{aligned}
\label{eq:lidar_polar}
\end{equation}
where $(r_j,\theta_j)$ are derived from the in-box points. When a Cartesian representation is required, the detection is mapped
to the world frame $\mathcal{W}$ using the 2D planar transform in~\eqref{eq:lidar_to_world}.



\label{sec:Calibration_and_Detection}

\section{Particle Filtering and Sensor Selection}

The particle filter approximates the posterior distribution $p(\mathbf{x}_k|\mathbf{z}_{1:k})$ using $N$ weighted particles $\{\mathbf{x}_k^{(i)}, w_k^{(i)}\}_{i=1}^N$. Each particle represents a hypothesis of the vessel's state. 


\subsection{Initialization}
\label{subsec:init}
The particle filter is initialized at the first fused time bin that contains at least one valid detection.
We select the initial position from the most recent measurement in the bin, prioritizing LiDAR when available, otherwise using the camera measurement mapped to $\mathcal{W}$. The initial state mean is
$\mathbf{x}_0=[p_x,\,p_y,\,0,\,0]^\top$, i.e., zero initial velocity.

Particles are drawn from the initial prior $p(\mathbf{x}_0)$ as a Gaussian distribution,
\begin{equation}
\mathbf{x}_0^{(i)} \sim \mathcal{N}(\mathbf{x}_0,\mathbf{P}_0), \qquad
w_0^{(i)}=\tfrac{1}{N},
\end{equation}
which follows the standard particle-filter assumption that an initial prior is available and can be sampled
\cite{arulampalam2002,barshalom_li_kirubarajan_2001}.

The initial covariance is chosen to reflect the uncertainty of the sensor used for initialization:
\begin{equation}
\mathbf{P}_0 =
\begin{bmatrix}
\boldsymbol{\Sigma}_{xy} & \mathbf{0}\\
\mathbf{0} & \sigma_{v0}^2\mathbf{I}_2
\end{bmatrix}.
\end{equation}
If initialization uses the camera, $\boldsymbol{\Sigma}_{xy}$ is obtained by propagating pixel uncertainty through the pixel→world mapping Jacobian and adding a calibration noise floor.
For LiDAR initialization we propagate $\boldsymbol{\Sigma}_{r,\theta}$  into $\boldsymbol{\Sigma}_{xy}$ using the Jacobian evaluated at the world-frame bearing. The initial velocity standard deviation $\sigma_{v0}$ is set to allow immediate motion after initialization.

\subsection{Prediction Step}
\label{subsec:prediction}
At each fused time bin $k$ with step $\Delta t_k$, each particle state
$\mathbf{x}^{(i)}_k=[p_x,\,p_y,\,v_x,\,v_y]^\top$ is propagated in two stages.
First, a deterministic kinematic model $f_{m_k}(\cdot)$ is applied:
\begin{equation}
\mathbf{x}^{(i)}_{k|k-1}=f_{m_k}\!\left(\mathbf{x}^{(i)}_{k-1},\,\Delta t_k\right).
\label{eq:pf_pred_deterministic}
\end{equation}
Second, process noise is injected using the discrete-time white-noise acceleration  model \cite{barshalom_li_kirubarajan_2001}:
\begin{equation}
\begin{split}
\mathbf{a}^{(i)}_k &\sim \mathcal{N}\!\left(\mathbf{0},\,\sigma_a^2\mathbf{I}_2\right),\\
\mathbf{v}^{(i)}_k &= \mathbf{v}^{(i)}_{k|k-1} + \Delta t_k\,\mathbf{a}^{(i)}_k,\\
\mathbf{p}^{(i)}_k &= \mathbf{p}^{(i)}_{k|k-1} + \tfrac{1}{2}\Delta t_k^2\,\mathbf{a}^{(i)}_k,
\end{split}
\label{eq:pf_pred_noise}
\end{equation}
where $\mathbf{p}=[p_x,p_y]^\top$ and $\mathbf{v}=[v_x,v_y]^\top$. We used $\sigma_a = 3~\mathrm{m/s^2}$ as a default value, which provides sufficient maneuverability for the vessel. 

\subsection{Update Step}
\label{subsec:update_step}
At each fused time bin $k$, the particle set $\{x^{(i)}_{k|k-1}, w^{(i)}_{k|k-1}\}_{i=1}^{N}$ is updated using the measurement set or sets available from the selected sensor configuration. We follow sequential importance sampling: for each selected sensor $s\in\mathcal{S}_k\subseteq\{\mathrm{cam},\mathrm{lidar}\}$ we compute a particle-wise likelihood $L_s(x^{(i)}_{k|k-1})$ and update weights multiplicatively,
\begin{equation}
\begin{aligned}
&w^{(i)}_{k} \propto
w^{(i)}_{k|k-1}
\prod_{s\in\mathcal{S}_k}
L_s\!\left(x^{(i)}_{k|k-1}\right), \\
&w^{(i)}_{k}\leftarrow
\frac{w^{(i)}_{k}}{\sum_{j=1}^{N} w^{(j)}_{k}} .
\end{aligned}
\label{eq:pf_weight_update}
\end{equation}
For numerical stability, this was implemented in the log-space \cite{arulampalam2002}. It should be highlighted that the current formulation does not introduce an additional explicit cross-sensor reliability coefficient between camera and LiDAR; instead, differences in sensor influence arise through the sensor-specific likelihood models, measurement uncertainties, and target-vs-clutter mixture terms defined below.

\subsubsection{Camera measurement-set likelihood}
For a camera frame at time bin $k$, the detector outputs a set of candidate world-frame measurements
$Z_{\mathrm{cam},k}=\{(z_j,R_j,c_j)\}_{j=1}^{M}$, where $z_j=[x_j,y_j]^\top$ is the mapped world position,
$R_j\in\mathbb{R}^{2\times 2}$ is the associated spatial uncertainty, and $c_j\in[0,1]$ is a heuristic detection-confidence proxy derived from the bounding-box area and foreground fill ratio and used to weight retained camera proposals within the frame.
For particle $x^{(i)}$, the predicted camera measurement is
\begin{equation}
h_{\mathrm{cam}}(x^{(i)}) =
\begin{bmatrix} p_x^{(i)} \\ p_y^{(i)} \end{bmatrix}.
\label{eq:h_cam}
\end{equation}

To avoid committing to a single association in clutter, we adopt a multi-hypothesis  update in which the
measurement-set likelihood is modeled as a mixture of target-origin and clutter-origin hypotheses:
\begin{equation}
\widetilde{L}_{\mathrm{cam}}(x^{(i)}) =
\alpha_{\mathrm{cam}}\,\ell_{\mathrm{cam}}(x^{(i)})
+ (1-\alpha_{\mathrm{cam}})\,\bar u_{\mathrm{cam}} .
\label{eq:cam_target_clutter_mixture}
\end{equation}
Here, $\ell_{\mathrm{cam}}(x^{(i)})$ is the target-origin term, defined as a confidence-weighted mixture of per-candidate
Gaussian likelihoods:
\begin{equation}
\begin{aligned}
&\ell_{\mathrm{cam}}(x^{(i)}) =
\sum_{j=1}^{M}\bar c_j\,
\mathcal{N}\!\big(z_j;\,h_{\mathrm{cam}}(x^{(i)}),\,R_j\big), \\
&\bar c_j = \frac{c_j}{\sum_{\ell=1}^{M} c_\ell}.
\end{aligned}
\label{eq:cam_target_term}
\end{equation}
The normalized weights $\bar c_j$ ensure that more reliable detections contribute more strongly while preserving a proper
convex mixture. The clutter-origin term $\bar u_{\mathrm{cam}}$ is a uniform spatial density over the assumed surveillance
region. The mixing coefficient $\alpha_{\mathrm{cam}}\in[0,1]$ captures the (set-level) probability that the current frame
contains a target-origin measurement (can be derived from detector confidence statistics and a nominal clutter rate, or tuned
as a robustness parameter), following the general sensor-management rationale in \cite{kyriakides2021agile}. 

\subsubsection{LiDAR measurement-set likelihood}
For each LiDAR scan, the detector outputs a set of range--bearing candidates
$Z_{\mathrm{lidar},k}=\{(z_j,c_j)\}_{j=1}^{M}$ where $z_j=[r_j,\theta^{w}_j]^\top$ uses a world-frame bearing
$\theta^{w}=\theta^{\mathrm{lidar}}+\psi$ (LiDAR$\rightarrow$world yaw $\psi$) and and $c_j\in[0,1]$ is a heuristic detection-confidence proxy calculated for the LiDAR scan in the similar way as for the camera frame: it is derived from the bounding-box area and foreground fill ratio and used to weight retained LiDAR proposals within the scan. For particle $x^{(i)}$, the predicted
range--bearing measurement is
\begin{equation}
h_{\mathrm{lidar}}(x^{(i)})=
\begin{bmatrix}
\sqrt{(p_x^{(i)}-s_x)^2+(p_y^{(i)}-s_y)^2}\\
\mathrm{atan2}(p_y^{(i)}-s_y,\;p_x^{(i)}-s_x)
\end{bmatrix},
\label{eq:h_lidar}
\end{equation}
with LiDAR position $[s_x,s_y]^\top$.

Analogously to (\ref{eq:cam_target_clutter_mixture})--(\ref{eq:cam_target_term}), we model measurement-origin uncertainty using a
target-vs-clutter mixture likelihood:
$$\widetilde{L}_{\mathrm{lidar}}(x^{(i)})=\alpha_{\mathrm{lidar}}\ell_{\mathrm{lidar}}(x^{(i)})+(1-\alpha_{\mathrm{lidar}})\bar u_{rb},$$
where $\ell_{\mathrm{lidar}}(\cdot)$ is formed as a confidence-weighted Gaussian mixture in $(r,\theta)$ space with a nominal
$R_{rb}=\mathrm{diag}(\sigma_r^2,\sigma_\theta^2)$, and $\bar u_{rb}$ is uniform on a declared range--bearing domain.


\subsection{Sensor Fusion Strategy}
\label{subsec:sensor_fusion_strategy}
At each fused time bin k, the filter applies the measurement likelihoods corresponding to the selected sensing regime. Depending on the regime, the update may use a single sensor likelihood or sequentially combine both available sensor likelihoods within the same bin. We therefore distinguish between measurement-level fusion (\emph{All sensors}) and sensor selection (\emph{Adaptive}). When multiple sensors are selected in the same bin, the implementation applies \emph{sequential}
updates (\ref{eq:pf_weight_update}). Under the standard conditional-independence assumption given the state, this is
equivalent to multiplying the selected likelihood factors (adding log-likelihoods). The filter supports four fusion configurations that determine which sensor likelihood factors are applied at each fused time bin.

\vspace{0.25em}
\subsubsection{LiDAR-only configuration}
If a LiDAR scan is available at bin $k$, set $\mathcal{S}_k=\{\mathrm{lidar}\}$ and update weights using the LiDAR
measurement-set likelihood. If LiDAR is
unavailable, no measurement update is performed in that bin.

\vspace{0.25em}
\subsubsection{Camera-only configuration}
\label{Camera-only regime section}
If a camera frame is available at bin $k$, set $\mathcal{S}_k=\{\mathrm{cam}\}$ and update weights using the camera
measurement-set likelihood. If the camera is unavailable, no measurement update is performed in that bin.

\vspace{0.25em}
\subsubsection{All sensors (camera + LiDAR)}
If all sensors are available at bin $k$, set $\mathcal{S}_k=\{\mathrm{cam},\mathrm{lidar}\}$ and apply sequential updates
(camera then LiDAR). If only one sensor is available, the update falls back to the available sensor only. This configuration corresponds to sequential measurement-level fusion of the available sensor likelihoods within the same fused time bin; it does not introduce an additional explicit cross-sensor weighting coefficient.

\vspace{0.25em}
\subsubsection{Adaptive configuration}
The filter selects the single most informative sensor at each bin using a entropy-reduction
criterion, following the general approach for information-driven sensing \cite{kyriakides2021agile, Xu_Gaussian_Process_Based_EIG}.
Let the predicted particle weights after the prediction step be $\mathbf{w}_{k|k-1}=\{w^{(i)}_{k|k-1}\}_{i=1}^{N}$ and define
the Shannon entropy:
\begin{equation}
H(\mathbf{w}_{k|k-1}) = -\sum_{i=1}^{N} w^{(i)}_{k|k-1}\log_2 w^{(i)}_{k|k-1}
\label{eq:entropy_weights}
\end{equation}

To select the most informative modality, we approximate the information gain that would be obtained by updating with each
candidate sensor $s\in\{\mathrm{cam},\mathrm{lidar}\}$ using a hypothetical measurement. Specifically, we treat the predicted estimate as if it was a detection $z^{\mathrm{hyp}}_{s,k}$ generated by sensor $s$. For each sensor $s$, a hypothetical update is then performed on a cloned particle set using $z^{\mathrm{hyp}}_{s,k}$ and nominal measurement-noise parameters, producing hypothetical posterior weights
$\mathbf{w}^{(s)}_{k}=\{w^{(i,s)}_{k}\}_{i=1}^{N}$ and the corresponding posterior entropy
\begin{equation}
H^{(s)}_{\mathrm{post}} = H(\mathbf{w}^{(s)}_{k})
\label{eq:H_post}
\end{equation}
The information gain score for sensor $s$ is defined as the entropy reduction
\begin{equation}
I_s = H_{\mathrm{prior}} - H^{(s)}_{\mathrm{post}},
\label{eq:IG_def}
\end{equation}
and the selected modality is $s_k^\star=\arg\max_s I_s$.
In the present implementation, the \emph{Adaptive} configuration selects a single modality from ${cam,lidar}$ at each fusion bin and should therefore be interpreted as an information-driven sensor-selection policy rather than an explicit dynamic reliability-weighted fusion rule.


\subsection{Resampling}
\label{subsec:resampling}
To prevent weight degeneracy, resampling is performed whenever the effective sample size, approximated as $N_{\text{eff}} \approx 1 / \sum_{i=1}^N (w_k^{(i)})^2$, falls below a threshold $N_{\text{thresh}}$. The systematic resampling algorithm is employed to replicate particles with high importance weights and eliminate those with negligible weights. Following resampling, the particle set is mapped to a uniform distribution, resetting all weights to $w_k^{(i)} = 1/N$.
\subsection{State Estimation}
\label{subsec:estimation}
The final state estimate is computed as:
\begin{equation}
\hat{\mathbf{x}}_k = \sum_{i=1}^N w_k^{(i)} \mathbf{x}_k^{(i)}
\end{equation}

\label{sec:Particle_Filtering_and_Sensor_Selection}

\section{Results and Analysis}
\label{sec:result_analysis}
\subsection{Experimental Setup and Evaluation Protocol}

\begin{figure}[!b]
\centering
\includegraphics[width=0.99\columnwidth]{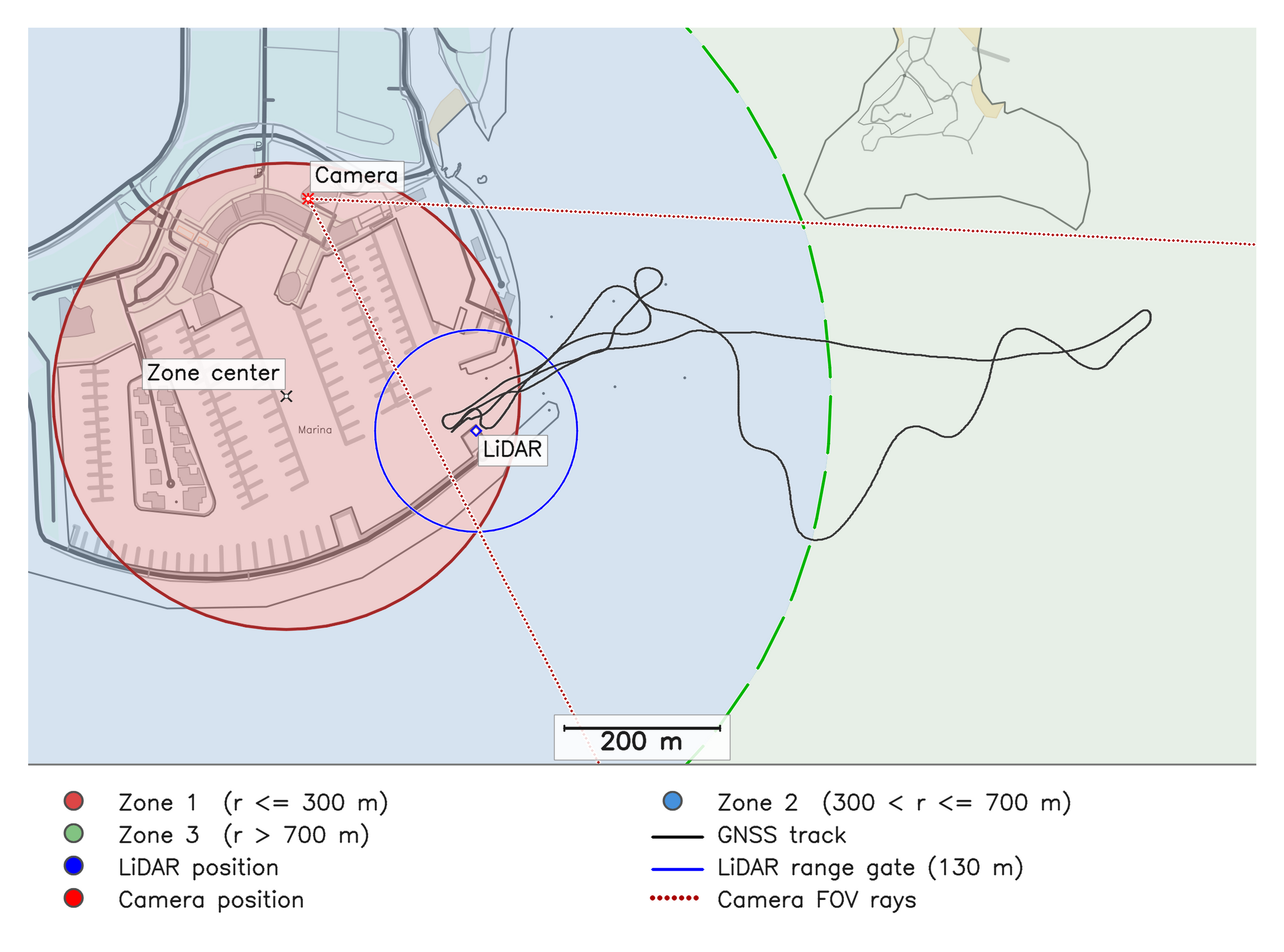}
\caption{Map context showing the three evaluation zones (Zone~1: marina, Zone~2: near, Zone~3: far) and the GNSS reference trajectory; zone assignment is based on the GNSS position.}
\label{fig:Map_zoning}
\end{figure}

Given the safety-critical nature of marina operations, performance was analyzed in a spatially stratified manner by partitioning the area of interest into three zones with increasing distance from the marina basin. As illustrated in Fig.~\ref{fig:Map_zoning}, Zone~1 (red) corresponds to the marina basin and immediate internal waters, where the most stringent requirements on tracking accuracy and continuity apply. Zone~2 (blue) covers the adjacent near-field approach region, which remains operationally important due to proximity to marina traffic and infrastructure. Zone~3 (green) represents the far-field region, where the objective is to maintain the best possible tracking performance while tolerating reduced sensing support.

All zones were defined as concentric annuli in the local world frame about a fixed center $\mathbf{c}$, with radii
$r_1=325$~m and $r_2=650$~m:
Zone~1 if $\|\mathbf{p}-\mathbf{c}\|\le r_1$,
Zone~2 if $r_1<\|\mathbf{p}-\mathbf{c}\|\le r_2$,
and Zone~3 if $\|\mathbf{p}-\mathbf{c}\|> r_2$.

A static sensor array comprising a shore-mounted Ouster OS2 LiDAR and an elevated fixed camera were used to track a rigid inflatable boat (RIB). A GNSS receiver onboard the RIB provided ground-truth positions. All measurements were mapped into a common local Cartesian ``world'' frame (meters) and time-aligned to a single reference clock. The evaluation timeline was formed by discretizing time into fixed-duration fusion bins of length $\Delta t_{\mathrm{bin}}$. Each time bin was assigned to a zone using the ground-truth position $\mathbf{p}^{\mathrm{GNSS}}_k$ to avoid bias from estimator drift. Within each bin, the most recent available measurement from each sensor was used to construct the fused measurement set. The particle filter (see used parameters in the Table~\ref{tab:pf_params}) produced a position estimate $\hat{\mathbf{p}}_k=[\hat{x}_k,\hat{y}_k]^{\mathsf{T}}$ at each bin $k$, which was compared against the corresponding ground-truth GNSS position $\mathbf{p}^{\mathrm{GNSS}}_k$.

\subsubsection{Metrics}
Let the per-bin planar position error be
\begin{equation}
e_k = \left\|\hat{\mathbf{p}}_k - \mathbf{p}^{\mathrm{GNSS}}_k\right\|_2 .
\label{eq:pos_err}
\end{equation}
For each zone $\Omega$, the position accuracy was summarized by the root-mean-square error (RMSE) over bins whose ground-truth fell in that zone:
\begin{equation}
\mathrm{RMSE}_{\Omega} = \sqrt{\frac{1}{|\mathcal{K}_{\Omega}|}\sum_{k\in\mathcal{K}_{\Omega}} e_k^2},
\label{eq:rmse_zone}
\end{equation}
where $|\mathcal{K}_{\Omega}|$ denotes the cardinality (size) of a set of bin indices assigned to zone $\Omega$.

Track continuity was quantified using a persistent \emph{lost-time-bin} labeling rule. A bin is considered \emph{bad} if
$e_k > e_{\mathrm{lost}}$ or $e_k$ is non-finite, where $e_{\mathrm{lost}}=50$~m; otherwise it is considered \emph{good}
($e_k \le e_{\mathrm{lost}}$). The track enters the \emph{lost} state only after $n=5$ consecutive bad bins, and exits
\emph{lost} only after $n=5$ consecutive good bins. At each transition, the last $n$ bins are retrospectively
marked/unmarked to avoid flickering.

The percentage of lost bins in zone $\Omega$ was computed as
\begin{equation}
\mathrm{Lost}\%_{\Omega} = 100\cdot \frac{1}{|\mathcal{K}_{\Omega}|}
\sum_{k\in\mathcal{K}_{\Omega}} \mathbb{I}^{\mathrm{lost}}_k ,
\label{eq:lost_pct}
\end{equation}
where $\mathbb{I}^{\mathrm{lost}}_k \in \{0,1\}$ denotes the final lost-bin label after applying the above persistence rule.

\subsubsection{Monte Carlo protocol}
Results were obtained via repeated Monte Carlo runs. For each $run\in\{1,\dots,N_{\mathrm{runs}}\}$, the four fusion configurations (\textit{LiDAR-only}, \textit{Camera-only}, \textit{All sensors}, and \textit{Adaptive}) were executed independently on the same time-aligned measurement timeline. Total number of runs is $N_{\mathrm{runs}} = 200$. To ensure repeatability and enable controlled comparisons, a deterministic random seed was assigned per run and re-used when initializing each regime within that run. For each $(run,configuration)$ pair, we computed zone-wise $\mathrm{RMSE}_\Omega$ and $\mathrm{Lost}\%_{\Omega}$; results were then aggregated across runs using mean value and confidence intervals (95\% bands) to characterize variability. The reported confidence intervals summarize stochastic variability across repeated executions of the same tracker on the same synchronized measurement sequence.

\begin{table}[t]
\caption{Particle-filter configuration used in the experiments.}
\label{tab:pf_params}
\centering
\footnotesize
\setlength{\tabcolsep}{4pt}          
\renewcommand{\arraystretch}{1.1}    
\begin{tabular}{l c}
\hline
\textbf{Parameter} & \textbf{Value} \\
\hline
Number of particles $N$ & 1000 \\
Resampling threshold $\tau$ (resample if $\mathrm{ESS}/N<\tau$) & 0.5 \\
Process noise (white-acceleration) $\sigma_a$ & $3~\mathrm{m/s^2}$ \\
LiDAR measurement noise $(\sigma_r,\sigma_\theta)$ & $(1~\mathrm{m},~0.5^\circ)$ \\
Camera centroid noise $\sigma_{\mathrm{px}}$ & $5~\mathrm{px}$ \\
LiDAR range gate $r_{\max}$ & $130~\mathrm{m}$ \\
Time-bin size $\Delta t_{\mathrm{bin}}$ & $0.2~\mathrm{s}$ \\
\hline
\end{tabular}
\end{table}

\subsection{Results description}
\begin{figure}[!b]
\centering
\includegraphics[width=0.99\columnwidth]{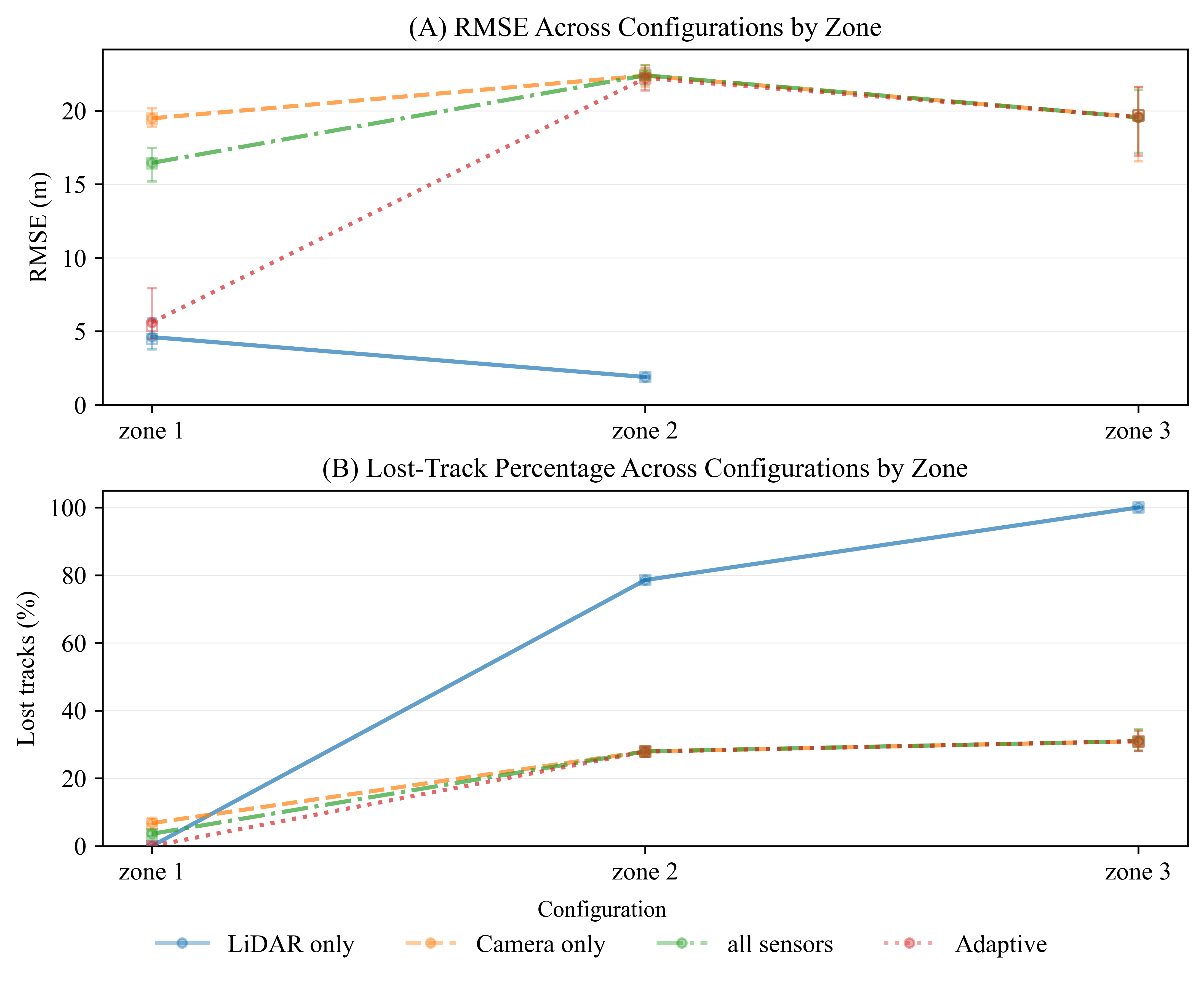}
\caption{Zone-wise tracking performance across fusion regimes: RMSE (A) between the particle-filter position estimate and GNSS ground truth; Lost-bin percentage by zone (B). Markers denote mean, error bars indicate 95\%  confidence intervals.}
\label{fig:RMSE_with_Lost_tracks_percentage}
\end{figure}

Let us outline at first the configuration-level behavior that is expected from the sensing geometry and measurement availability. Fig.~\ref{fig:RMSE_with_Lost_tracks_percentage} and Table~\ref{tab:zone_wise_tracking_performance} report zone-wise accuracy and track continuity  for the four configurations (\textit{LiDAR-only}, \textit{Camera-only}, \textit{All sensors}, \textit{Adaptive}), using mean RMSE with 95\% confidence intervals and the lost-time-bin percentage.  Due to the imposed LiDAR range gating, LiDAR detections are predominantly available in Zone~1 and only marginally in Zone~2 (see Fig.~\ref{fig:Map_zoning}). Consequently, \textit{LiDAR-only} configuration is expected to provide the best performance in Zone~1, whereas camera-driven configuration is expected to dominate in Zones~2--3 where LiDAR measurements are absent. The \textit{Camera-only} configuration provides coverage across all zones but typically exhibits higher position uncertainty due to projection and detection noise. The \textit{All sensors} configuration applies all available measurements; in Zone~1 it can benefit from complementary constraints, although its performance may be limited by the higher-noise camera likelihood relative to LiDAR. Finally, the \textit{Adaptive} configuration is expected to select LiDAR when informative/available (primarily Zone~1) and to revert to \textit{Camera-only} updates when LiDAR becomes unavailable (Zones~2--3). 

Fig.~\ref{fig:RMSE_with_Lost_tracks_percentage} compares the four configurations discussed above.
In Zone 1, the \emph{LiDAR-only} configuration attains the lowest RMSE and a low lost time-bin fraction, which is consistent with strong near-field geometric constraints and the availability of LiDAR detections within the imposed range gate (Fig.~\ref{fig:Map_zoning}). The \emph{Adaptive} configuration achieves comparable performance in this zone because the information-driven policy predominantly selects LiDAR when it is the most informative modality. The fact that the All sensors configuration does not outperform \emph{LiDAR-only} indicates that, in the near field, sequential incorporation of the camera likelihood does not necessarily improve the estimate when camera-based localization is subject to larger projection and detection uncertainty. This behavior should therefore be interpreted as a limitation of unweighted joint likelihood updates, and it motivates the use of modality selection when sensor informativeness is strongly regime-dependent.

\begin{table*}[!t]
\centering
\caption{Zone-wise tracking performance corresponding to Fig.~\ref{fig:RMSE_with_Lost_tracks_percentage}. RMSE values are reported in meters, and lost-bin values are reported as percentages.}
\label{tab:zone_wise_tracking_performance}
\footnotesize
\setlength{\tabcolsep}{4pt}
\renewcommand{\arraystretch}{1.15}
\begin{tabular}{llcccccc}
\toprule
\multirow{2}{*}{Configuration} & \multirow{2}{*}{Zone}
& \multicolumn{3}{c}{RMSE (m)}
& \multicolumn{3}{c}{Lost bins (\%)} \\
\cmidrule(lr){3-5} \cmidrule(lr){6-8}
& & Mean & Median & 95\% CI & Mean & Median & 95\% CI \\
\midrule
LiDAR-only
& Zone 1 & 4.67 & 4.52 & [3.84, 6.13] & 0.00 & 0.00 & [0.00, 0.00] \\
& Zone 2 & 1.89 & 1.89 & [1.85, 1.94] & 78.56 & 78.58 & [78.40, 78.58] \\
& Zone 3 & -- & -- & -- & 100.00 & 100.00 & [100.00, 100.00] \\
\midrule
Camera-only
& Zone 1 & 19.50 & 19.50 & [18.94, 20.17] & 6.82 & 6.42 & [6.31, 8.45] \\
& Zone 2 & 22.36 & 22.38 & [21.58, 23.12] & 27.93 & 27.95 & [26.42, 29.25] \\
& Zone 3 & 19.60 & 19.70 & [16.99, 21.78] & 30.80 & 30.71 & [27.85, 34.45] \\
\midrule
All sensors
& Zone 1 & 16.43 & 16.40 & [14.79, 17.48] & 3.59 & 3.54 & [2.90, 4.61] \\
& Zone 2 & 22.39 & 22.36 & [21.68, 23.20] & 27.93 & 27.90 & [26.42, 29.25] \\
& Zone 3 & 19.60 & 19.66 & [17.17, 21.76] & 30.83 & 30.71 & [27.99, 34.51] \\
\midrule
Adaptive
& Zone 1 & 5.64 & 5.36 & [4.52, 8.14] & 0.00 & 0.00 & [0.00, 0.00] \\
& Zone 2 & 22.20 & 22.19 & [21.45, 22.92] & 27.93 & 27.99 & [26.60, 29.16] \\
& Zone 3 & 19.60 & 19.67 & [17.20, 21.62] & 30.96 & 30.84 & [28.26, 33.97] \\
\bottomrule
\end{tabular}
\end{table*}

In Zone~2, the \emph{Camera-only}, \emph{All sensors}, and \emph{Adaptive} configurations show similar RMSE levels and similar
lost time-bin percentages, indicating that camera updates largely govern performance once LiDAR detections become intermittent.
Although \emph{LiDAR-only} may yield low RMSE on the subset of time bins where tracking is maintained, its continuity degrades
substantially (approximately $80\%$ of time bins are labeled LOST), consistent with operation near the effective LiDAR limit.
Accordingly, RMSE values for \emph{LiDAR-only} in Zone~2 should be interpreted jointly with the lost time-bin metric.

In Zone~3, \emph{LiDAR-only} becomes unreliable (lost time-bin percentage approaches $100\%$), and RMSE is not reported due to
insufficient valid time bins. By comparison, \emph{Camera-only}, \emph{All sensors}, and \emph{Adaptive} maintain nonzero
continuity but with elevated RMSE and non-negligible lost time-bin fractions, consistent with long-range operation where
camera-based localization dominates and uncertainty increases.

Thus, across zones, the \emph{Adaptive} configuration approximates the best fixed configuration in each operating region by selecting LiDAR in the near field and reverting to camera updates when LiDAR detections are unavailable, yielding a favorable accuracy--continuity trade-off over the full trajectory.

It should be noted that beyond accuracy and continuity, the adaptive configuration offers an operational benefit: it can reduce sensing-system load by selecting \emph{one} modality per fusion bin, instead of continuously processing all streams~\cite{hero2008infotheory}. In practical deployments, such selection can translate to \emph{retasking}: a non-selected sensor can be allocated to other objectives, such as tracking additional targets, higher-rate classification, or recording for forensic review, aligning with resource allocation viewpoints in managed sensing~\cite{hero2008infotheory}. The same
principle can provide lower compute and bandwidth usage by avoiding full-rate processing of all modalities.
Accordingly, Fig.~\ref{fig:RMSE_with_Lost_tracks_percentage} compares the relative behavior of \emph{LiDAR-only}, \emph{Camera-only}, \emph{All sensors}, and \emph{Adaptive} configurations under a common estimation approach.



\section{Conclusion}
\label{sec:conclusion}
Experimental evaluation at Ayia Napa Marina with onboard GNSS ground truth highlighted the operating trade-offs among the considered sensing regimes. LiDAR-only tracking achieved the highest accuracy in the near field (Zone 1), consistent with strong geometric constraints when LiDAR detections are available, while camera-driven updates maintained coverage as the target moved beyond the effective LiDAR range. Importantly, these results provide \emph{real-world} experimental evidence that an information-gain  (entropy-reduction) based sensor-selection strategy can be effective in a realistic maritime environment, extending prior studies that have largely emphasized simulation-based evaluation~\cite{kyriakides2021agile}. The reported results therefore support a practical combination of two mechanisms: sequential measurement-level fusion when both modalities are used, and information-driven sensor selection when one modality is preferred. In particular, the Adaptive regime should be interpreted as a sensor-selection policy that favors the more informative modality at each fusion bin, rather than as an explicit dynamic reliability-weighted fusion rule.


Beyond estimation performance, the adaptive configuration can be interpreted as a lightweight sensor-management mechanism:
by activating only the modality expected to be most informative at each fusion bin, it can reduce computational and
communication load and enable \emph{retasking} of non-selected sensors to other objectives (e.g., additional targets,
classification, or archival recording). These benefits depend on platform-level support for selective throttling
(e.g., duty-cycling or reduced-rate processing) and are not unconditional: aggressive single-sensor selection reduces
redundancy and can increase sensitivity to occlusions or modality-specific failure modes. Accordingly, practical
deployments should incorporate conservative fallback rules (e.g., temporarily reverting to multi-sensor updates when
uncertainty increases)~\cite{hero2011sensormanagement}.

The proposed approach is modular and constitutes a baseline that will be extended to richer heterogeneous sensing suites, including multiple cameras (RGB/IR/thermal), passive acoustic arrays, marine radar, and laser-based sensors, enabling fusion across fundamentally different physical modalities. Heterogeneous multi-sensor maritime tracking has been widely motivated
to improve robustness under varying operating conditions (including visibility and weather effects)~\cite{helgesen2022asv_heterogeneous}. Although the evaluation is limited to a single recorded real-world trajectory from one deployment site, the sequence is operationally diverse, containing multiple marina entry and exit maneuvers as well as strongly non-linear motion with sharp turns. The results should therefore be interpreted as a case study demonstrating operating-regime behavior, rather than as a statistically broad benchmark. Broader validation across multiple trajectories, vessel types, and environmental conditions remains future work. Also, future work will transition from single-target tracking to multi-target tracking with principled data association, and will improve robustness to adverse weather/sea-state and long-range operation via enhanced detection models, range-dependent uncertainty calibration, and more expressive motion models. Also, an important next step is to replace the current unweighted joint update with an explicit reliability-aware fusion mechanism, for example via state-dependent uncertainty scaling or sensor-specific weighting of the measurement likelihoods. Another major next step is comparison against conventional multi-sensor tracking baselines, such as Extended Kalman Filter, Unscented Kalman Filter, and Ensemble Kalman Filter based fusion and track-level fusion methods, in order to position the proposed particle-filter approach more broadly against alternative estimator families. Finally, automated synchronization and online calibration
updates would reduce reliance on manual anchor events and improve deployability in long-duration real-world
surveillance~\cite{xiong2022_online_temporal_calib,persic2021_online_multisensor_calib}.

\section*{Acknowledgment}
The authors thank the Cyprus Marine and Maritime Institute (CMMI) for access to the Smart Marina Testbed at Ayia Napa
Marina and the associated sensing equipment. The testbed infrastructure was made available through the MDigi-I project
``Maritime Digitalization Research Infrastructure'' (STRATEGIC INFRASTRUCTURES/1222/0113), implemented under the Cohesion
Policy Programme ``THALIA 2021--2027'' and co-funded by the Republic of Cyprus and the European Regional Development Fund (ERDF).

This work was supported by the LORELEI-X project (Grant Agreement No. 101159489), funded by the European Union under the Horizon Europe Programme.

The authors would also like to thank Michael Picas and Kyriacos Clerides of the Cyprus Marine and Maritime Institute (CMMI) for their technical support during the experimental setup, sensor deployment, and data collection activities.

CMMI was established as a “Center of Excellence” in Marine and Maritime Research, Technology Development \& Innovation (RTDI) and has received funding from the European Union’s Horizon 2020 research and innovation program under grant agreement No. 857586 and matching funding from the Government of the Republic of Cyprus.
\bibliographystyle{IEEEtran}
\bibliography{references}

\end{document}